\documentclass[letterpaper, 10 pt, conference]{ieeeconf}  

\IEEEoverridecommandlockouts                              

\usepackage{amsmath,amsfonts}
\usepackage{array}
\usepackage[caption=false,font=normalsize,labelfont=sf,textfont=sf]{subfig}
\usepackage{textcomp}
\usepackage{stfloats}
\usepackage{url}
\usepackage{verbatim}
\usepackage{graphicx}
\usepackage{cite}
\usepackage{multirow}
\usepackage{lscape} 
\usepackage{graphicx} 
\usepackage{comment}
\usepackage{color}
\usepackage{algorithm}
\usepackage[noend]{algpseudocode} 
\usepackage{amssymb,amsmath,comment,cite}
\usepackage{xspace}
\usepackage{tabularray}

\newtheorem{definition}{Definition}

\usepackage{balance}
\usepackage{subcaption}

\usepackage{booktabs}
\usepackage{array}

\newcommand\MOPP{MOPP\xspace}

\begin{document}
	
\title{\LARGE \bf
PWTO: A Heuristic Approach for Trajectory Optimization \\ in Complex Terrains
}



\author{Yilin Cai, Zhongqiang Ren
	\thanks{Yilin Cai is with Georgia Institute of Technology, Atlanta, GA 30332, USA. 
	(email: yilincai@gatech.edu). Zhongqiang Ren is with Shanghai Jiao Tong University, Shanghai, 200240, China.(email: zhongqiang.ren@sjtu.edu.cn)}%
}

	\maketitle

	
	
	\begin{abstract}
		This paper considers a trajectory planning problem for a robot navigating complex terrains, which arises in applications ranging from autonomous mining vehicles to planetary rovers. The problem seeks to find a low-cost dynamically feasible trajectory for the robot. The problem is challenging as it requires solving a non-linear optimization problem that often has many local minima due to the complex terrain. To address the challenge, we propose a method called Pareto-optimal Warm-started Trajectory Optimization (PWTO) that attempts to combine the benefits of graph search and trajectory optimization, two very different approaches to planning. PWTO first creates a state lattice using simplified dynamics of the robot and leverages a multi-objective graph search method to obtain a set of paths. Each of the paths is then used to warm-start a local trajectory optimization process, so that different local minima are explored to find a globally low-cost solution. In our tests, the solution cost computed by PWTO is often less than half of the costs computed by the baselines. In addition, we verify the trajectories generated by PWTO in Gazebo simulation in complex terrains with both wheeled and quadruped robots. The code of this paper is open sourced and can be found at \url{https://github.com/rap-lab-org/public_pwto}.

	\end{abstract}

	\section{Introduction}\label{pwdc:sec:intro}
	
Optimal trajectory planning in complex terrains is of fundamental importance in robotics.
This problem arises in applications ranging from planetary rovers~\cite{strader2020perception}, exploration~\cite{Fan-RSS-21} to autonomous mining~\cite{berglund2009planning}.
The problem is challenging as it requires solving a non-linear optimization problem to find a low-cost trajectory in complex terrains while ensuring dynamic-feasibility along the trajectory.
A common approach to solve this problem is to first generate a geometric path free of kinematic and dynamic constraints, and then use that path to warm-start a trajectory optimization process which ultimately respects the dynamics of the robot \cite{andreasson2015fast,Fan-RSS-21}.
This type of approach has two major limitations:
first, it can lead to a highly sub-optimal solution that converges to a local minimum;
second, the local trajectory optimization may require many iterations to converge due to the complicated objective function and the dynamic constraints of the robot.

\begin{figure}[t]
	\centering
	\includegraphics[width=\linewidth]{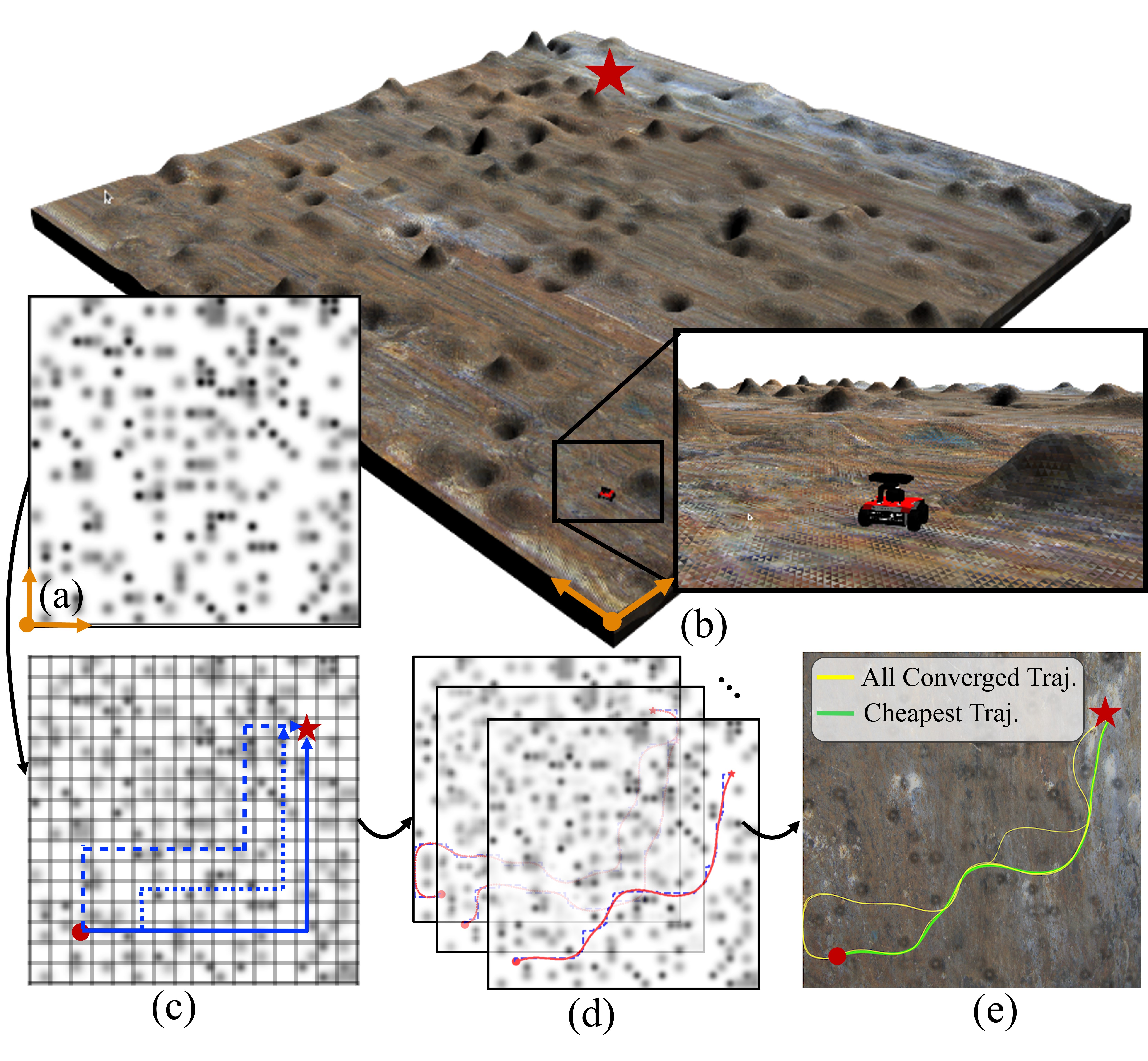}
	\caption{
		Overview of proposed Pareto-optimal Warm-started Trajectory Optimization method for robot trajectory planning to navigate complex terrains. (a-b) A continuous traversability cost field and its corresponding terrain map. (c) A set of Pareto-optimal paths obtained from multi-objective graph search in the state lattice. (d) Trajectories reported by PWTO in a round-robin fashion. (e) Visualization of the trajectories for robot to track on rough terrain.
	}
	\label{pwdc:fig:fig1}
\end{figure}

This paper develops a new heuristic approach to handle the problem by combining graph search and trajectory optimization.
By using concepts such as Pareto-optimality from multi-objective optimization, we propose a method called Pareto-optimal Warm-started Trajectory Optimization (PWTO).
PWTO (Fig.~\ref{pwdc:fig:fig1}) first creates a state lattice embedded in a low-dimensional state space by using simplified dynamics of the robot, and leverages a multi-objective graph search method to obtain a set of Pareto-optimal paths in the lattice.\footnote{A solution path is Pareto-optimal if there exists no other solution that will yield an improvement in one objective without causing a deterioration in at least one of the other objectives.}
Each of the paths then seeds a local trajectory optimization process, which explores different local minima and is more likely to find a globally low-cost solution.
PWTO runs these optimization processes in a round-robin fashion and reports the converged solution in an anytime fashion. The converged trajectory solutions allow the robot to track them and navigate through complex terrains.



We evaluate PWTO in various terrains, against baselines where different types of initial guesses are used.
Although PWTO has no solution optimality guarantees in theory, the results show that, in practice, PWTO often computes solutions with less than half of the cost of the baselines, and is able to avoid being trapped in an optimization process that requires many iterations to converge.
Our Gazebo simulation verifies that the planned trajectory can be executed in several different terrains.


	\section{Related Work}\label{pwdc:sec:related}
	
\subsection{Terrain Traversability}
Robot navigation in complex terrains is an important research topic that involves traversability analysis, trajectory planning, etc.
Terrain traversability is a broad concept with varying meaning in different applications~\cite{papadakis2013terrain}.
Building a traversability map is out of the scope of this article, and we limit our focus to a trajectory planning problem for a dynamic robot within a 2D cost field that indicates the terrain traversal cost.
Different from a discrete occupancy grid that is commonly used to describe obstacles for conventional mobile robot planning, the cost field in this work is continuous, and often involves many local minima, which makes the trajectory optimization a challenging problem.

\subsection{Dynamically Feasible Trajectory Planning}
Planning dynamically feasible trajectories for a mobile robot has been widely studied.
Sampling-based methods (e.g. \cite{karaman2011sampling}) can quickly return a dynamically feasible trajectory and keep refining the solution quality thereafter.
However, the solution obtained is often far away from the optimum and the refinement may converge slowly without domain-specific fine tuning.
Search-based methods such as~\cite{pivtoraiko2011kinodynamic} can provide high-quality solutions with global theoretic guarantees by using a set of pre-computed dynamically feasible motion primitives.
However, to ensure the dynamic feasibility, a high-dimensional search space that captures the full state of the robot is often required, and search-based methods are hard to scale to high-dimensional search spaces in general.
In contrast, optimization-based methods such as~\cite{zucker2013chomp} scale well in high-dimensional state space by iteratively running local optimization but often rely on a good initial guess to warm-start the optimization in order to bypass local minima.

Hybrid approaches that combine the benefits of sampling, search and optimization have also been investigated, and this paper focuses on methods that combine search and optimization.
There are mainly two strategies to combine search and optimization.
First, optimization can be embedded into an A*-like search as a procedure to generate motion primitives (i.e., a short dynamically feasible trajectory that connect two adjacent states of A* search).
Existing methods that follow this strategy (such as~\cite{natarajan2021interleaving}) typically consider a binary representation of the workspace (i.e., either free or occupied by obstacles) and is thus different from the continuous cost field considered in this work.
The second strategy is to use an A* search to find a path in a graph that is embedded in some low-dimensional space, and then use the path to seed the trajectory optimization.
Methods following this strategy~\cite{andreasson2015fast,Fan-RSS-21} have been applied to similar problems and our approach belongs to this category.

\subsection{Multi-Objective Path Planning (\MOPP)}
This paper is also inspired by the recent advance in \MOPP.
Given a graph where each edge is associated with a non-negative cost vector (where each component of the vector corresponds to an objective), \MOPP aims to find a set of Pareto-optimal (also called non-dominated) paths connecting the given start and destination vertex in the graph.
\MOPP has been studied for decades~\cite{loui1983optimal} and remains an active research area~\cite{ren2022enhanced,HernandezYBZSKS23,ren22mopbd,ren22mosipp}.
Conventionally, this problem has been regarded as computationally expensive especially when there are many Pareto-optimal paths.
The recent advances in Multi-Objective A* (MOA*) search~\cite{ren2022enhanced,HernandezYBZSKS23} expedite the search and make it possible to employ MOA* as a sub-routine to solve more challenging planning problems.
This work is an attempt along this direction.

	\section{Problem Description}\label{pwdc:sec:problem}
	
Let $\mathcal{W}\subset SE(2)$ denote a bounded workspace.
In this paper, we use three real numbers $p_x,p_y,\theta$ to represent $SE(2)$.
Let $C:\mathcal{W}\rightarrow \mathbb{R}$ denote a twice differentiable potential field defined over the workspace.
Let $\dot{\mathbf{x}}(t)=f(\mathbf{x}(t),\mathbf{u}(t)), t\geq0$ denote the dynamics of the robot, where $x$ is a $n$-dimensional state ($n\geq 2$) of the robot and $u$ is a $m$-dimensional control of the robot.
For any state $\mathbf{x}$, let $\mathbf{x}_p\in\mathcal{W}$ denote the pose of the robot contained in state $\mathbf{x}$.
 
For $\mathbf{x}(t),\mathbf{u}(t)$, let
\begin{eqnarray}\label{pwdc:eqn:obj_J}
J(\mathbf{x}(t),\mathbf{u}(t),T) = \int_{t=0}^{T} C \left( \mathbf{x}_p(t) \right) + \mathbf{u}(t)^T \mathbf{R} \mathbf{u}(t) \text{d}t
\end{eqnarray}
denote the cost of the trajectory, where $\mathbf{R}$ is a positive semi-definite matrix.
Intuitively, $J$ is the sum of both the cost incurred by the potential field and the control effort within the horizon of the trajectory.
This paper considers the following trajectory optimization problem:
\begin{eqnarray}
&\min_{\mathbf{x}(t),\mathbf{u}(t),T} J(\mathbf{x}(t),\mathbf{u}(t),T) \\
&\text{s.t.}\;\;\; \dot{\mathbf{x}}(t) = f(\mathbf{x}(t),\mathbf{u}(t),T) , \; t \geq 0\\
&\mathbf{x}(0) = \mathbf{x}_{init},\; \mathbf{x}(T) = \mathbf{x}_{goal},\; \mathbf{x}(t) \in \mathcal{X},\; \mathbf{u}(t) \in \mathcal{U}
\end{eqnarray}
where $\mathcal{X}$ and $\mathcal{U}$ denote the eligible set of states and controls of the robot respectively at any time during the robot motion, $\mathbf{x}_{init}$ and $\mathbf{x}_{goal}$ denote the initial and goal states of the robot respectively.
In other words, the problem seeks to find a dynamically feasible trajectory from $\mathbf{x}_{init}$ to $\mathbf{x}_{goal}$, while the trajectory cost $J$ is minimized.

	
	\section{Method}\label{pwdc:sec:method}
	
\subsection{Overview}
Our method PWTO is shown in Alg.~\ref{pwdc:alg:pwdc}, which consists of two parts.
Part I (Lines 1-3) relaxes the original trajectory optimization problem into a multi-objective path planning (\MOPP) problem over a graph by simplifying the dynamics of the robot.
The multi-objective is formulated based on the following observations: (i) the original optimization problem (\ref{pwdc:eqn:obj_J}) inherently involves optimizing multiple criteria, such as the trajectory horizon $T$, the cost incurred by the potential field, etc, where each criterion affects the objective $J$; (ii) by discretizing the state and control spaces using simplified dynamics, a corresponding \MOPP problem can be formulated and efficiently solved by leveraging the recent \MOPP algorithms to obtain a set of Pareto-optimal paths.

Part II (Lines 4-14) uses each of the Pareto-optimal paths computed in the first part to seed a corresponding \emph{trajectory optimization process}, which is hereafter referred to as a \emph{process} to simplify the presentation.
Due to the complicated cost field $C$, it is often hard to predict how many iterations each process takes before convergence.
If these processes were optimized till convergence one after another (i.e., in a sequential manner), the method could return the first solution trajectory after many optimization iterations, since the first process may take a huge number of iterations to converge.
To bypass this issue, PWTO chooses to optimize these processes in a round-robin (i.e. episodic) manner, where each process is optimized for $K$ (i.e., a user-defined constant) iterations within an episode.
PWTO terminates after a maximum number of episodes (denoted as $N_{episode}$) defined by the user.
This yields an \emph{anytime} algorithm that can potentially output the first solution quickly during the computation, and as the runtime increases, more processes may converge and the min-cost solution is reported during the computation.

\begin{algorithm}[tbp]
	\caption{PWTO}\label{pwdc:alg:pwdc}
 \small
	\begin{algorithmic}[1]
		\State{$(G,\vec{c},v_{init},v_{goal})$ $\gets$ \textit{GenerateMOPP}()}
		\State{$\Pi_*$ $\gets$ \textit{SolveMOPP}($G,\vec{c},v_{init},v_{goal}$)}
		\State{$\Pi'$ $\gets$ \textit{Filter}($\Pi_*$)}
		\State{$\mathcal{P}$ $\gets$ \textit{InitOptProcesses}($\Pi'$)} \Comment{A set of optimization processes}
		\State{$n_{episode}$ $\gets$ 0, $C_{min}$ $\gets \infty$ }
		\While{$n_{episode} < N_{episode}$}
		\ForAll{$p \in \mathcal{P}$}
		\State{run $p$ for $K$ iterations} 
		\If{$p$ converges}
		\State{remove $p$ from $\mathcal{P}$}
		  \If{$p.cost < C_{min}$}
		  \State{$C_{min} \gets p.cost$}
		  \State{report $p$}
          \EndIf		
		\EndIf
		\EndFor
		\State{$n_{episode} \gets n_{episode} + 1$}
		\EndWhile
		\State{\textbf{return}}
	\end{algorithmic}
\end{algorithm}

\subsection{Part I: Multi-Objective Path Planning}\label{pwdc:sec:method:mospp}
\subsubsection{State and Action Spaces}
The state space $\mathcal{X}$ of the robot is 5-dimensional.
Instead of directly running graph search in $\mathcal{X}$, PWTO first considers only the pose of the robot in the workspace and ignores the linear and angular velocities by searching in a low-dimensional state space.
Specifically, PWTO discretizes the workspace $\mathcal{W}$ into a state lattice $G=(V,E)$, where each vertex $v\in V$ represents a possible pose of the robot in $SE(2)$, and each edge $e \in E$ in the lattice corresponds to a pre-computed motion primitive that moves the robot from one vertex to another.
To compute the motion primitives, a simplified dynamics (i.e., first-order dynamics) of the robot $(\dot{p_x},\dot{p_y},\dot{\theta}) = (v\,cos(\theta),v\,sin(\theta),\omega), \mathbf{u}=[v,\omega]^T$ is used.
For each state $\mathbf{x} \in \mathcal{X}$, let $v(x)$ denote a unique nearest vertex to $\mathbf{x}_p$ in the lattice using an arbitrary measure in $SE(2)$. Ties can be broken by choosing the vertex with the smallest coordinates.
Additionally, let $v_{init}:= v(x_{init,p})$ and $v_{goal}:= v(x_{goal,p})$.

\subsubsection{Vector-cost Edges}\label{pwdc:sec:method:vec_edge_cost}
For each edge $e\in E$, a non-negative cost vector $\vec{c}(e)$ is assigned to represent the cost with respect to different criteria.
We consider the following two types of cost for each edge.
The first cost (i.e., ${c}_1(e)$) is the traversal time of the motion primitives.
The second cost is incurred by the cost field $C$, and for each edge $e=(u,v)$, the second cost is the integral of the corresponding motion primitive trajectory in the cost field. 
Each edge $e\in E$ is now associated with a two-dimensional non-negative cost vector $\vec{c}(e) = (c_1(e),c_2(e))$.
Note that, other choices of edge costs are also possible to be used within PWTO.

\subsubsection{Non-nominated Paths}
Let $\pi(v_1,v_\ell)=(v_1,v_2,\dots,v_\ell)$ denote a path of length $\ell$, which is a sequence of vertices in $V$ where any pair of adjacent vertices $(v_k,v_{k+1}),k=1,2,\dots \ell-1$ are connected by an edge in $E$. The cost of $\pi(v_1,v_\ell)$ is the sum of edge costs that are present in the path (i.e., $\vec{c}(\pi(v_1,v_\ell)):= \sum_{k=1}^{\ell-1} \vec{c}(v_k,v_{k+1})$).
To compare two paths, their corresponding cost vectors are compared using the dominance relation~\cite{ehrgott2005multicriteria}.
\begin{definition}[Dominance]\label{def:dominance}
	Given two vectors $\vec{a}$ and $\vec{b}$ of length $M$, $\vec{a}$ dominates $\vec{b}$ (denoted as $\vec{a} \succeq \vec{b}$) if and only if $\vec{a}(m) \leq \vec{b}(m)$, $\forall m \in \{1,2,\dots,M\}$, and $\vec{a}(m) < \vec{b}(m)$, $\exists m\in \{1,2,\dots,M\}$.
\end{definition}
Any two paths $\pi_1(u,v),\pi_2(u,v)$ that connect two vertices $u,v \in V$ are non-dominated by each other, if the corresponding cost vectors do not dominate each other.
Among all possible paths from $v_{init}$ to $v_{goal}$, the set of non-dominated paths is called the Pareto-optimal set, whose corresponding cost vectors are called the Pareto-optimal front.

\subsubsection{MOA* Search}
Given the aforementioned state lattice $G$, the edge cost vectors $\vec{c}$, and the initial and goal vertices $v_{init},v_{goal}$, the set of Pareto-optimal paths can be found by using MOA* algorithms.\footnote{The MOA* algorithm used in this work finds a set of ``cost-unique'' Pareto-optimal paths. In other words, if two paths have the same cost vector, only one of them is found. MOA* algorithms that can find all Pareto-optimal paths can also be used within the PWTO approach (Line 2 in Alg.~\ref{pwdc:alg:pwdc}).}
This paper uses our prior Enhanced Multi-Objective A* (EMOA*)~\cite{ren2022enhanced} algorithm to find the entire Pareto-optimal front as well as a corresponding set of cost-unique Pareto-optimal paths $\Pi_*$.

\subsubsection{Filtering Paths Using Hausdorff Distance}
The solution path set $\Pi_*$ may contain many Pareto-optimal paths, and using all of them to seed the trajectory optimization can lead to a large number of processes.
Many paths in $\Pi_*$ have similar shapes to each other, although they have different cost vectors.
We thus introduce another comparison method to post-process (i.e., filter) $\Pi_*$ in order to obtain a subset of paths that have different shapes from each other.


We treat each path $\pi \in \Pi_*$ as a set of points in $\mathbb{R}^2$ and compare two paths using the Hausdorff distance~\cite{schutze2012using}.
Here, Hausdorff distance is just one way to compare the similarity of two paths, and other approaches (such as homotopy classes) can also be used.
\vspace{1mm}
\begin{definition}[Hausdorff Distance]\label{def:hausdorff}
	Given any two paths $\pi_1$ and $\pi_2$ in $G$, the Hausdorff distance is $$d_H(\pi_1,\pi_2) := \max\{\sup_{v_1\in \pi_1} d_\pi(v_1,\pi_2), \sup_{v_2\in \pi_2} d_\pi(v_2,\pi_1)\},$$
\end{definition}
\vspace{1mm}
where $d_\pi(v,\pi) := \inf_{v'\in \pi}d(v,v')$ and $d(v,v')$ is the Euclidean distance between points $v,v'$.
Intuitively, a large $d_H(\pi_1,\pi_2)$ indicates that $\pi_1,\pi_2$ are distinct from each other.

To filter paths in $\Pi_*$, we introduce a hyper-parameter $d_{H,thres}>0$, and iterate all pairs of distinct paths $\pi_1,\pi_2 \in \Pi_*$ to compute a subset of paths $\Pi'$ such that the Hausdorff distance between any pair of paths in $\Pi'$ is greater than $d_{H,thres}$ (i.e., $d_H(\pi_i,\pi_j) > d_{H,thres} \forall \pi_i,\pi_j \in \Pi', i\neq j$).
Note that $\Pi'$ is not unique, and here we find an arbitrary $\Pi'$.
Then, the filtered path set $\Pi'$ will be used to warm-start the trajectory optimization.
Here, parameter $d_{H,thres}$ is specified by the user before the computation starts, and a larger $d_{H,thres}$ leads to a smaller set $\Pi'$.


\begin{table*}[!b]
\centering
\caption{Comparison between the solution costs of 10 instances computed by PWTO and the baselines in various maps. Comparison is evaluated as the solution cost computed by the baseline divided by the solution cost computed by PWTO.}
\label{tab:result}
\begin{tabular}{@{}cccccccccccccccccc@{}}
\toprule
\multicolumn{1}{c}{\multirow{2}{*}{\textbf{Index}}} & \multirow{2}{*}{\textbf{Cost Field}}                                                                  & \multicolumn{2}{c|}{\textbf{Parameters}}                                    & \multicolumn{3}{c|}{\textbf{Line/PWTO}}                                                                                               & \multicolumn{3}{c|}{\textbf{Random/PWTO}}     & \multicolumn{3}{c|}{\textbf{t-RRT/PWTO}}                                                                                                 & \multicolumn{3}{c|}{\textbf{A*/PWTO}}                                                                                                    & \multicolumn{2}{c}{\textbf{Total}}                  \\ \cmidrule(l){3-18} 
\multicolumn{1}{c}{}                                &                                                                                                       & \multicolumn{1}{c}{$\sigma_m$}             & \multicolumn{1}{c|}{M}                   & \multicolumn{1}{c}{$>$1}                  & \multicolumn{1}{c}{$>$2}                  & \multicolumn{1}{c|}{Fail}               & \multicolumn{1}{c}{$>$1}                  & \multicolumn{1}{c}{$>$2}                  & \multicolumn{1}{c|}{Fail}               & \multicolumn{1}{c}{$>$1}                  & \multicolumn{1}{c}{$>$2}                  & \multicolumn{1}{c|}{Fail}                  & \multicolumn{1}{c}{$>$1}                  & \multicolumn{1}{c}{$>$2}                  & \multicolumn{1}{c|}{Fail}                  & \multicolumn{1}{c}{$>$1} & \multicolumn{1}{c}{$>$2} \\ \midrule
\multirow{2}{*}{1}                                  & \multirow{2}{*}{\includegraphics[width=0.8cm]{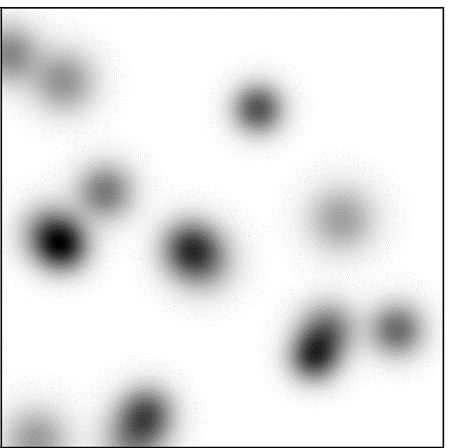}} & \multicolumn{1}{c}{\multirow{2}{*}{0.002}} & \multicolumn{1}{c|}{\multirow{2}{*}{15}} & \multicolumn{1}{c}{\multirow{2}{*}{1.00}} & \multicolumn{1}{c}{\multirow{2}{*}{0.22}} & \multicolumn{1}{c|}{\multirow{2}{*}{0}} & \multicolumn{1}{c}{\multirow{2}{*}{0.78}} & \multicolumn{1}{c}{\multirow{2}{*}{0.33}} & \multicolumn{1}{c|}{\multirow{2}{*}{0}} & \multicolumn{1}{c}{\multirow{2}{*}{1.00}} & \multicolumn{1}{c}{\multirow{2}{*}{0.44}} & \multicolumn{1}{c|}{\multirow{2}{*}{0.33}} & \multicolumn{1}{c}{\multirow{2}{*}{1.00}} & \multicolumn{1}{c}{\multirow{2}{*}{0.78}} & \multicolumn{1}{c|}{\multirow{2}{*}{0.33}} & \multirow{2}{*}{0.94}    & \multirow{2}{*}{0.44}    \\
                                                    &                                                                                                       & \multicolumn{1}{c}{}                       & \multicolumn{1}{c|}{}                    & \multicolumn{1}{c}{}                      & \multicolumn{1}{c}{}                      & \multicolumn{1}{c|}{}                   & \multicolumn{1}{c}{}                      & \multicolumn{1}{c}{}                      & \multicolumn{1}{c|}{}                   & \multicolumn{1}{c}{}                      & \multicolumn{1}{c}{}                      & \multicolumn{1}{c|}{}                      & \multicolumn{1}{c}{}                      & \multicolumn{1}{c}{}                      & \multicolumn{1}{c|}{}                      &                          &                          \\
\multirow{2}{*}{2}                                  & \multirow{2}{*}{\includegraphics[width=0.8cm]{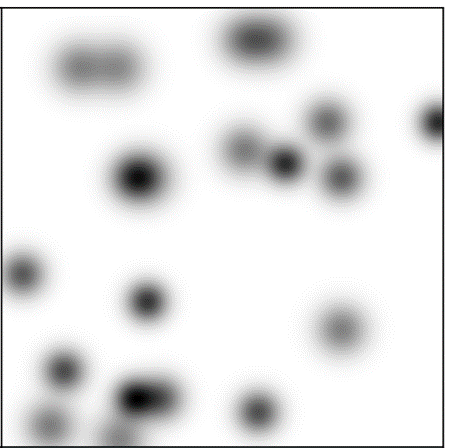}} & \multirow{2}{*}{0.012}                     & \multicolumn{1}{l|}{\multirow{2}{*}{20}} & \multirow{2}{*}{0.67}                     & \multirow{2}{*}{0.33}                     & \multicolumn{1}{l|}{\multirow{2}{*}{0}} & \multirow{2}{*}{0.67}                     & \multirow{2}{*}{0.22}                     & \multicolumn{1}{l|}{\multirow{2}{*}{0}} & \multirow{2}{*}{1.00}                     & \multirow{2}{*}{0.89}                     & \multicolumn{1}{l|}{\multirow{2}{*}{0.56}} & \multirow{2}{*}{1.00}                     & \multirow{2}{*}{0.67}                     & \multicolumn{1}{l|}{\multirow{2}{*}{0.44}} & \multirow{2}{*}{0.83}    & \multirow{2}{*}{0.53}    \\
                                                    &                                                                                                       &                                            & \multicolumn{1}{l|}{}                    &                                           &                                           & \multicolumn{1}{l|}{}                   &                                           &                                           & \multicolumn{1}{l|}{}                   &                                           &                                           & \multicolumn{1}{l|}{}                      &                                           &                                           & \multicolumn{1}{l|}{}                      &                          &                          \\
\multirow{2}{*}{3}                                  & \multirow{2}{*}{\includegraphics[width=0.8cm]{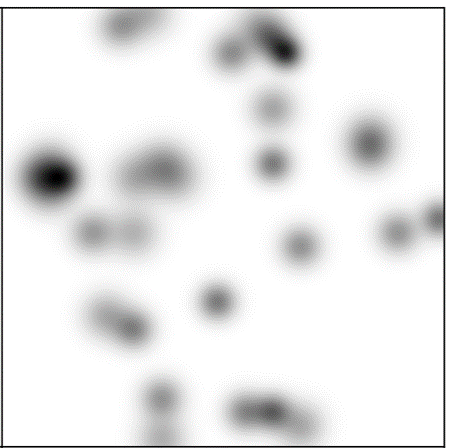}} & \multirow{2}{*}{0.001}                     & \multicolumn{1}{l|}{\multirow{2}{*}{30}} & \multirow{2}{*}{0.60}                     & \multirow{2}{*}{0.40}                     & \multicolumn{1}{l|}{\multirow{2}{*}{0}} & \multirow{2}{*}{0.70}                     & \multirow{2}{*}{0.50}                     & \multicolumn{1}{l|}{\multirow{2}{*}{0}} & \multirow{2}{*}{0.90}                     & \multirow{2}{*}{0.70}                     & \multicolumn{1}{l|}{\multirow{2}{*}{0.50}} & \multirow{2}{*}{1.00}                     & \multirow{2}{*}{0.90}                     & \multicolumn{1}{l|}{\multirow{2}{*}{0.40}} & \multirow{2}{*}{0.80}    & \multirow{2}{*}{0.63}    \\
                                                    &                                                                                                       &                                            & \multicolumn{1}{l|}{}                    &                                           &                                           & \multicolumn{1}{l|}{}                   &                                           &                                           & \multicolumn{1}{l|}{}                   &                                           &                                           & \multicolumn{1}{l|}{}                      &                                           &                                           & \multicolumn{1}{l|}{}                      &                          &                          \\
\multirow{2}{*}{4}                                  & \multirow{2}{*}{\includegraphics[width=0.8cm]{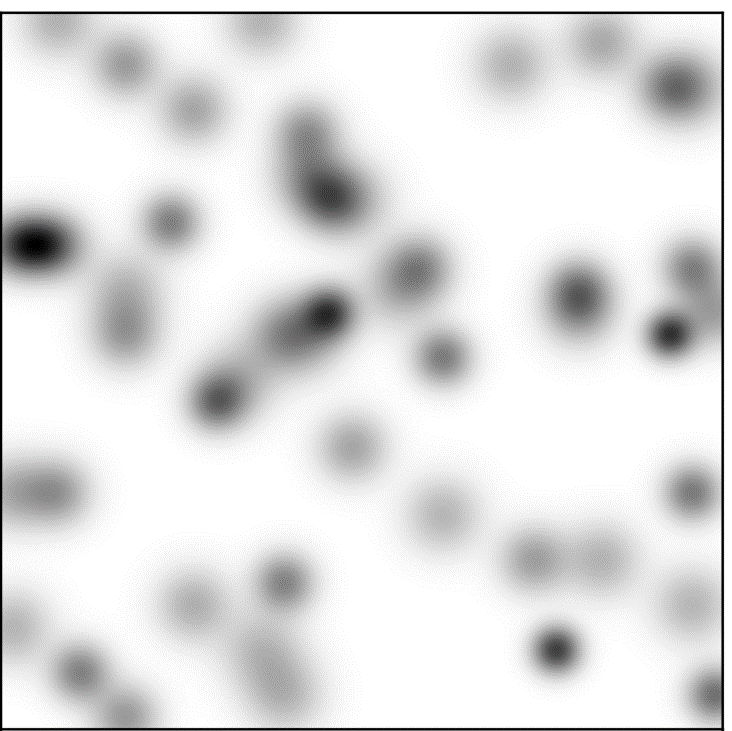}} & \multirow{2}{*}{0.0005}                    & \multicolumn{1}{l|}{\multirow{2}{*}{50}} & \multirow{2}{*}{1.00}                     & \multirow{2}{*}{0.44}                     & \multicolumn{1}{l|}{\multirow{2}{*}{0}} & \multirow{2}{*}{1.00}                     & \multirow{2}{*}{0.44}                     & \multicolumn{1}{l|}{\multirow{2}{*}{0}} & \multirow{2}{*}{1.00}                     & \multirow{2}{*}{0.44}                     & \multicolumn{1}{l|}{\multirow{2}{*}{0.67}} & \multirow{2}{*}{1.00}                     & \multirow{2}{*}{0.56}                     & \multicolumn{1}{l|}{\multirow{2}{*}{0.33}} & \multirow{2}{*}{1.00}    & \multirow{2}{*}{0.58}    \\
                                                    &                                                                                                       &                                            & \multicolumn{1}{l|}{}                    &                                           &                                           & \multicolumn{1}{l|}{}                   &                                           &                                           & \multicolumn{1}{l|}{}                   &                                           &                                           & \multicolumn{1}{l|}{}                      &                                           &                                           & \multicolumn{1}{l|}{}                      &                          &                          \\ \bottomrule
\end{tabular}
\end{table*}

\subsection{Part II: Trajectory Optimization}\label{pwdc:sec:method:trajOpt}
The goal of the trajectory optimization is to find a low-cost dynamically feasible trajectory near the initial guess.
Since the dynamics of the robot is simplified in the previous graph search step, the returned paths in $\Pi'$ are dynamically infeasible in general with respect to the original problem.
Therefore, the trajectory optimization method is required to be able to start the optimization from some dynamically infeasible initial guess, and we use the direct collocation (DirCol) algorithm~\cite{hargraves1987direct,underactuated}, a trajectory optimization method that accepts dynamically infeasible initial guesses.
Note that other methods that accept dynamically infeasible initial guesses can also be used in PWTO (line 8 in Alg.~\ref{pwdc:alg:pwdc}).

\subsubsection{Direct Collocation}
We use each path in the filtered path set $\Pi'$ to seed a trajectory optimization process and there are total $|\Pi'|$ number of processes (line 4 in Alg.~\ref{pwdc:alg:pwdc}).
Specifically, we first convert a path $\pi \in \Pi'$ into a state-control trajectory $\mathbf{x}_0(t),\mathbf{u}_0(t)$ by (i) interpolating between each pair of adjacent vertices in $\pi$ and (ii) taking finite differences to compute the additional terms in the states and controls (i.e., velocities and accelerations).
Then, $\mathbf{x}_0(t),\mathbf{u}_0(t)$ are used as the initial guess to warm-start a DirCol process.
In short, DirCol parameterizes the state trajectory and control trajectory as polynomials for each pair of adjacent time steps and transcribes the original continuous problem into a non-linear programming problem (NLP). Moreover, the robot dynamics is enforced as equality constraints at the middle point (i.e., collocation points) between each pair of adjacent time steps.
The resulting NLP is then solved by the existing NLP solver.
At convergence (when the change in $J'$ is smaller than a certain threshold), the resulting trajectory is dynamically feasible and the objective function is (locally) minimized.\footnote{DirCol computes the gradient and the Hessian matrix of the objective function with respect to the $x$ and $u$.
This paper thus requires that the cost field $C$ is twice differentiable in Sec.~\ref{pwdc:sec:problem}, so that the Hessian matrix can be computed for DirCol optimization.
This twice differentiable requirement can be discarded if the trajectory optimizer does not require Hessian.
}

\subsubsection{Reference Tracking Cost}
Due to the non-linearity of the NLP and the dynamically infeasible initial guess, using DirCol to directly optimize the $J$ in~(\ref{pwdc:eqn:obj_J}) may lead to trajectories that are trapped in local minima while going through high cost regions.
In other words, although the initial guess $\mathbf{x}_0(t),\mathbf{u}_0(t)$ avoids the high cost region in $C$, during the optimization of the NLP, to ensure the dynamic constraints, the trajectory may deviate a lot from the initial guess, go through high cost regions, and get trapped in a highly sub-optimal local minimum.

To avoid this, we introduce an additional \emph{reference tracking cost} (which is commonly used in optimal control to track a given reference) into the objective function of the NLP.
The intuition is to make the optimized trajectory stay close to the given initial guess, since the initial guess bypasses high cost regions in the cost field $C$.
Formally, let $J'$ denote the objective of the NLP problem, which differs from the original objective function defined in the problem statement by adding an additional term:
\begin{eqnarray}\label{pwdc:eqn:Jprime}
J' := J + \int_{t=0}^{T}(\mathbf{x}(t) - \mathbf{x}_0(t))^T \mathbf{Q} (\mathbf{x}(t) - \mathbf{x}_0(t)) \text{d}t
\end{eqnarray}
where $J$ is defined in~(\ref{pwdc:eqn:obj_J}), $\mathbf{Q}$ is a positive semi-definite matrix, and the added term helps the trajectory to stay close to the initial guess when the dynamic constraints are enforced during the optimization.

\subsubsection{Round-Robin Optimization}\label{pwdc:sec:method:episodic_optm}
PWTO runs the optimization processes in a round-robin (i.e., episodic) fashion (lines 6-14 in Alg.~\ref{pwdc:alg:pwdc}).
In each episode, the un-converged processes, which are stored in a set $\mathcal{P}$, are iterated and each process is optimized for $K$ iterations.
In each episode, the converged process is removed from $\mathcal{P}$, while the un-converged processes stays in $\mathcal{P}$ and will be optimized in the next episode.
The converged solutions are reported during the PWTO.
This episodic optimization allows PWTO to run in an \emph{anytime} fashion in a sense that: the first feasible solution can be obtained quickly while better solutions can be obtained as the runtime increases.

	\section{Numerical Results}\label{pwdc:sec:result}
	

\subsection{Implementation and Test Settings}

\subsubsection{Implementation and Baselines}
We implement our PWTO (Alg.~\ref{pwdc:alg:pwdc}) in Python, while leveraging a C++ implementation of EMOA*~\cite{ren2022enhanced} (Line 2 in Alg.~\ref{pwdc:alg:pwdc}), and a Python implementation of the DirCol~\cite{moore2018opty} (Line 8 in Alg.~\ref{pwdc:alg:pwdc}).
This DirCol implementation uses IPOPT~\cite{wachter2006implementation} as the NLP solver.\footnote{The C++ implementation of EMOA* is at \url{https://github.com/wonderren/public_emoa}. The Python implementation of DirCol is at \url{https://opty.readthedocs.io/en/latest/}. The IPOPT solver is at \url{https://coin-or.github.io/Ipopt/index.html}.}
All baselines warm-starts DirCol using different initial guesses.
The \textbf{first} baseline uses an initial guess that is a straight line connecting the start and goal position, and then linearly interpolates the intermediate states (denoted as ``LINE'').
The \textbf{second} baseline uses a randomly generated initial guess (denoted as ``RAND'').
The \textbf{third} baseline first leverages A* to find a path in a graph, and then uses the path to warm-start a DirCol process (denoted as ``A*'').
Here, the A* search is conducted in a graph $G=(V,E)$ with edge costs defined as $c(e) := 0.5c_1(e)+ 0.5c_2(e), \forall e\in E$, where $c_1,c_2$ are defined in the Method section.
The \textbf{fourth} baseline seeds DirCol using the path found by T-RRT~\cite{jaillet2008transition}, a sampling based algorithm that solves a similar problem to this paper.

\subsubsection{Dynamics of the Robot}
As aforementioned, in all tests, we use a second-order uni-cycle model.
We consider the following constraints: $(x,y)\in[0,1]\times[0,1]$ (stay in the workspace), $v \in [0,0.05]$, $\omega \in [-1.57,1.57]$ (limits on linear and angular velocities) and $a_v \in [-0.1,0.1]$, $a_w \in [-1,1]$ (limits on linear and angular acceleration).
For numerical computation, the dynamics is integrated using the forward Euler integration with a time interval of $0.1$ seconds.

\subsubsection{Cost Field Generation}
The $(p_x,p_y)$ coordinates of the cost fields are limited to $[0,1]\times[0,1]$.
We randomly sample $M$ Gaussian distributions $\mathcal{N}(\mu_m,\Sigma_m), m=1,2,\dots,M$, where $\mu_m$ is randomly sampled from $[0,1]\times[0,1]$, and $\Sigma_m:=[\sigma_m,0;0,\sigma_m]$ with $\sigma_m$ being a positive number randomly sampled from a certain range that is elaborated later.
The cost field is defined as the sum of these $M$ Gaussian distributions.
We generate three different cost fields as shown in Table~\ref{tab:result}, where the second row shows the parameters.
For MOPP, we discretize the cost field uniformly into a lattice of size $200\times200\times4$ which correspond to the number of discretized $x,y,\theta$ coordinates respectively.

\subsubsection{Other Parameters}
For each cost field, we generate 10 instances (i.e., start-goal pairs).
Each baseline is allowed to run at most 1000 optimization iterations per instance.
PWTO is allowed to run for 10 episodes per instance, where each episode has 100 optimization iterations.
To filter Pareto-optimal paths, we set $d_{H,thres}=8$ for all the tests.

\begin{figure}[!bt]
	\centering
	\includegraphics[width=\linewidth]{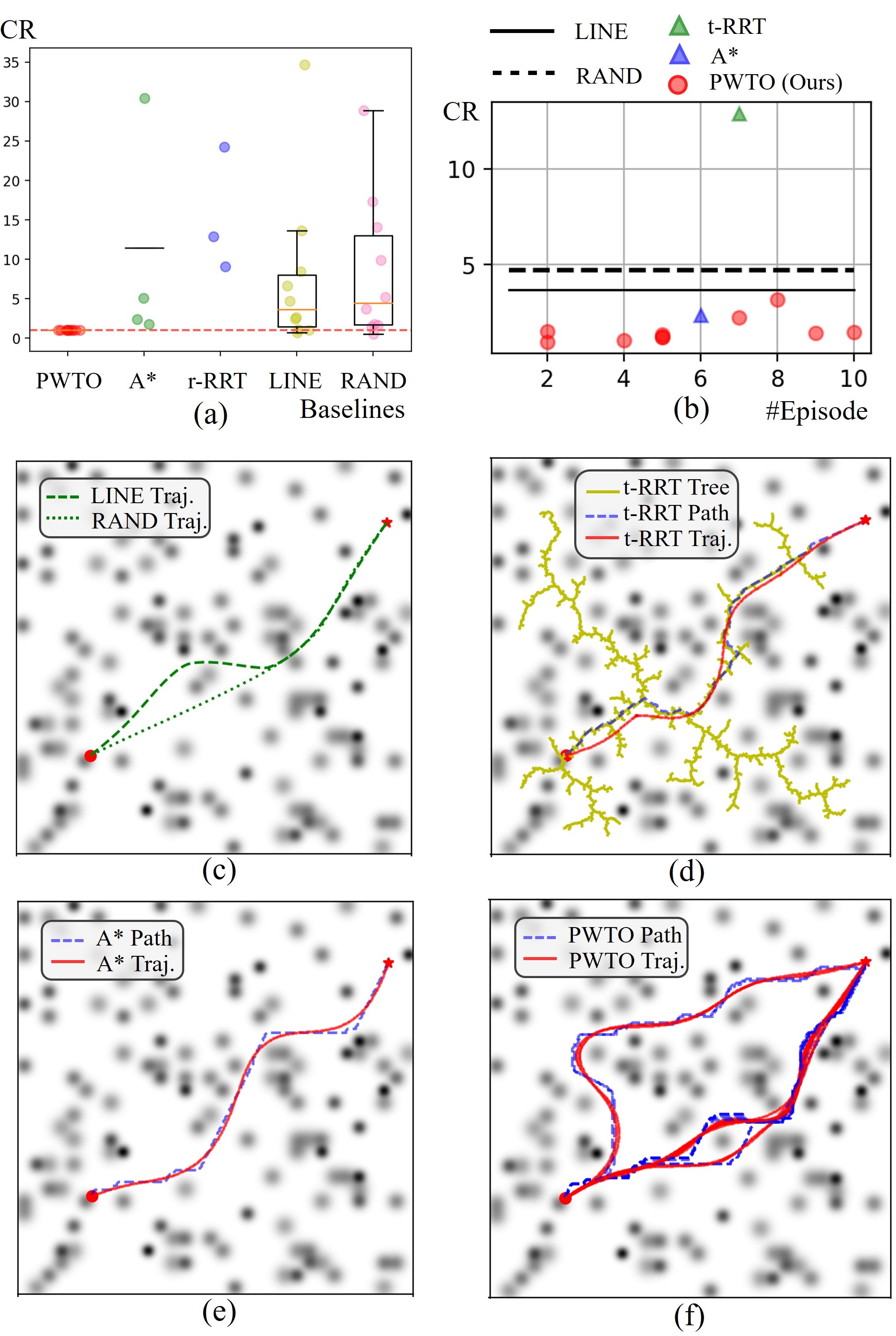}
  \vspace{-6mm}
	\caption{Test results of 10 instances in a more complicated cost field with $\sigma=0.0002, M=133$. The starting position is indicated as red circle and the goal is indicated as red star. (a) shows the CR of each baselines with failed instances excluded. (b-f) show detailed information about an instance where all approaches successfully solve. In (b), the horizontal axis shows the number of episodes required by PWTO to find each converged solution trajectory, and the vertical axis shows the cost ratio of each solution over the cheapest trajectory cost found by PWTO.
	(c-f) show the solution trajectories returned by each approach.
	}
 \vspace{-5mm}
	\label{pwdc:fig:instance_detail}
\end{figure}

\subsection{Trajectory Cost Comparison}


We compare the cost of the solutions computed by our PWTO and the baselines in various maps.
Let cost ratio (CR) denote the solution cost computed by a baseline divided by the solution cost computed by PWTO.
In other words, a CR that is greater than one indicates PWTO computes a cheaper solution than the baseline, and PWTO is more advantageous over the baselines when CR is larger.
We conduct numerical tests with 10 instances (sets of initial and goal state) for the maps and compute CR only for those instances where PWTO converges. For each map, there is at most one instance where PWTO fails to converge within the limits of optimization iterations.
As shown in Table \ref{tab:result}, PWTO often computes cheaper trajectories than the baselines for most of the instances.
Additionally, the solution cost of PWTO is less than half of the solution cost computed by the baselines (i.e., CR$>2$) for some instances.


We then create a more complicated cost field (Fig.~\ref{pwdc:fig:instance_detail} (c-f)) with 10 start-goal pairs and plot the CR of each baseline against PWTO in Fig.~\ref{pwdc:fig:instance_detail} (a). PWTO often computes cheaper trajectories than the other approaches.
We then pick an instance where all approaches successfully find a solution.
As shown in Fig.~\ref{pwdc:fig:instance_detail} (b), due to the round-robin optimization, PWTO can report the computed trajectories in an anytime fashion, and avoid spending too many iterations in any one specific optimization process.


\subsection{Simulation for Mobile Robots}

\begin{figure}[tb]
	\centering
	\includegraphics[width=\linewidth]{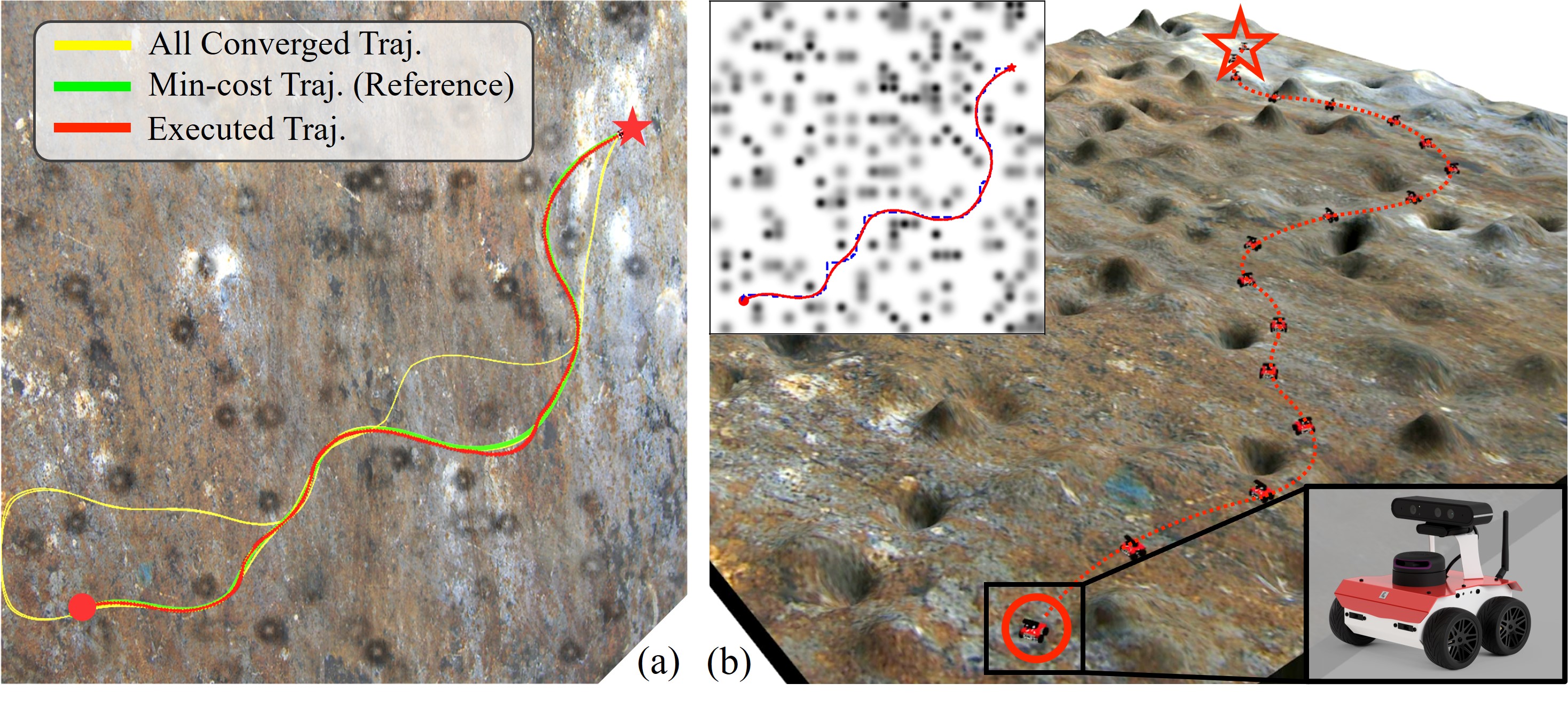}
	\caption{ Simulation and validation of the PWTO for the trajectory planning of mobile robots. (a) shows all the trajectories (yellow) reported during the PWTO computation, the cheapest trajectory (green) that is used as the reference, and the actual trajectory executed by the robot (red). (b) visualizes the Gazebo simulation. With the closed-loop control, the robot is able to follow the reference trajectory in complex terrains.}
	\label{pwdc:fig:sim_rosbot}
\end{figure}

To verify the dynamic feasibility of the computed trajectory, we use Gazebo and ROS to simulate the trajectory execution process of the robot in complex terrains.
We first generate the Gazebo terrain using the gazebo world construction tool provided  in~\cite{abbyasov2020automatic}, which takes in put of a gray-scale image of the cost field.
For visualization purposes, we use the absolute value of the height of the terrain to indicate the traversal cost.
For example, in Fig.~\ref{pwdc:fig:sim_rosbot}, flat land indicates small traversal cost while dents or hills indicate high traversal cost.\footnote{We acknowledge that roughness of the terrain is not directly related to the height of the terrain. Here, we use height for easy visualization.}
Then we use the ROSbot (HUSARION) platform to test all the trajectories.

To address the motion disturbance in the Gazebo simulation (e.g. caused by the friction between the wheel and the ground), we implement a feedback control law as described in~\cite{kanayama1990stable}.
Specifically, during the execution, the controller iteratively obtains the actual position of the robot from Gazebo and computes an error state as well as the feedback control term $u_{fb}$.
Note that the planned (reference) trajectory by PWTO is a state-control trajectory which includes the feedforward terms $u_{ff}$ (i.e. the acceleration of the robot at each time step) for execution.
The resulting control command is $u =u_{ff}+u_{fb}$, which is then bounded by the control limit of the robot and then send to the robot for execution.
Fig.~\ref{pwdc:fig:sim_rosbot} visualizes the resulting trajectory using the closed-loop control.
We tested the dynamic feasibility of the planned trajectory for the mobile under two different terrains with different complexity. 
More video demonstration of trajectory following tests can be found in our multi-media attachment. In short, the robot can follow the planned trajectory in complex terrains and reach the goal position successfully.



\subsection{Simulation for Quadruped Robots}

\begin{figure}[tb]
	\centering
	\includegraphics[width=\linewidth]{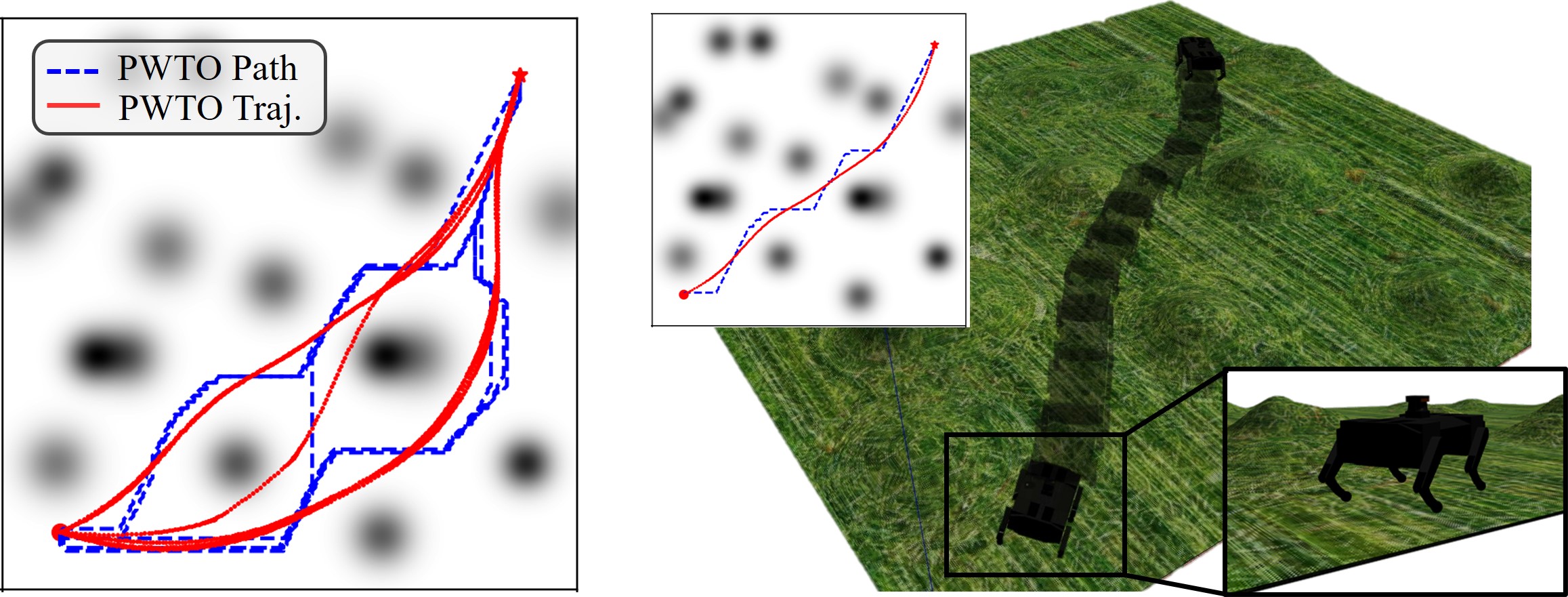}
	\caption{Simulation and validation of the PWTO for a quadruped robot. (a) shows all the paths (blue) and converged trajectories (red) reported during the PWTO computation. (b) visualizes the Gazebo simulation of a quadruped robot following the reference trajectory in complex terrains.}
 \vspace{-2mm}
	\label{pwdc:fig:sim_quad}
\end{figure}

Finally, we evaluate PWTO with a more complex quadruped robot.
Here, we take a hierarchical approach, where the center of mass of the quadruped robot is first planned by PWTO, which finds a trajectory for the robot to follow in the complex terrain. Then, to track the planned trajectory, we used an open source controller CHAMP\footnote{\url{https://github.com/chvmp/champ} }, which achieves highly dynamic locomotion utilizing pattern modulation and impedance control~\cite{lee2013hierarchical}.
By leveraging the CHAMP controller, we can directly sent the robot linear and angular velocity ($\dot{\mathbf{p}}$ , $\boldsymbol{\omega}_b$) as the command to the quadruped robot. The foot placement and swing command will be output by the CHAMP controller for low-level joint control.
The cost field here is similar to the aforementioned sum of Gaussian distributions with $M=10$ and $\sigma_m=0.003$.
The results shows that the proposed PTWO can be applied to various robot platforms and generate dynamical feasible trajectories for them to travel through the complex terrains. Further video demonstrations can be found in the multi-media attachment.

	\section{Conclusions and Future Work}\label{pwdc:sec:conclude}
	This paper considers a trajectory planning problem for a dynamic mobile robot in complex terrains.
We propose a heuristic method PWTO to solve the problem.
PWTO leverages multi-objective search techniques to seed multiple optimization processes, so that different local minima are explored to help find a globally low-cost trajectory.
In the meanwhile, PWTO runs these optimization processes in a round-robin fashion and is able to report the converged solution in an anytime fashion.
Future work includes theoretic proof on solution quality that guarantees PWTO to be able to find trajectories with cheaper costs than the baselines.

    	\bibliographystyle{IEEEtran}
	\bibliography{ref}

\begin{thebibliography}{10}
\providecommand{\url}[1]{#1}
\csname url@samestyle\endcsname
\providecommand{\newblock}{\relax}
\providecommand{\bibinfo}[2]{#2}
\providecommand{\BIBentrySTDinterwordspacing}{\spaceskip=0pt\relax}
\providecommand{\BIBentryALTinterwordstretchfactor}{4}
\providecommand{\BIBentryALTinterwordspacing}{\spaceskip=\fontdimen2\font plus
\BIBentryALTinterwordstretchfactor\fontdimen3\font minus \fontdimen4\font\relax}
\providecommand{\BIBforeignlanguage}[2]{{%
\expandafter\ifx\csname l@#1\endcsname\relax
\typeout{** WARNING: IEEEtran.bst: No hyphenation pattern has been}%
\typeout{** loaded for the language `#1'. Using the pattern for}%
\typeout{** the default language instead.}%
\else
\language=\csname l@#1\endcsname
\fi
#2}}
\providecommand{\BIBdecl}{\relax}
\BIBdecl

\bibitem{strader2020perception}
J.~Strader, K.~Otsu, and A.-a. Agha-mohammadi, ``Perception-aware autonomous mast motion planning for planetary exploration rovers,'' \emph{Journal of Field Robotics}, vol.~37, no.~5, pp. 812--829, 2020.

\bibitem{Fan-RSS-21}
D.~D. Fan, K.~Otsu, Y.~Kubo, A.~Dixit, J.~Burdick, and A.~akbar Agha-mohammadi, ``{STEP: Stochastic Traversability Evaluation and Planning for Risk-Aware Off-road Navigation},'' in \emph{Proceedings of Robotics: Science and Systems}, Virtual, July 2021.

\bibitem{berglund2009planning}
T.~Berglund, A.~Brodnik, H.~Jonsson, M.~Staffanson, and I.~Soderkvist, ``Planning smooth and obstacle-avoiding b-spline paths for autonomous mining vehicles,'' \emph{IEEE transactions on automation science and engineering}, vol.~7, no.~1, pp. 167--172, 2009.

\bibitem{andreasson2015fast}
H.~Andreasson, J.~Saarinen, M.~Cirillo, T.~Stoyanov, and A.~J. Lilienthal, ``Fast, continuous state path smoothing to improve navigation accuracy,'' in \emph{2015 IEEE International Conference on Robotics and Automation (ICRA)}.\hskip 1em plus 0.5em minus 0.4em\relax IEEE, 2015, pp. 662--669.

\bibitem{papadakis2013terrain}
P.~Papadakis, ``Terrain traversability analysis methods for unmanned ground vehicles: A survey,'' \emph{Engineering Applications of Artificial Intelligence}, vol.~26, no.~4, pp. 1373--1385, 2013.

\bibitem{karaman2011sampling}
S.~Karaman and E.~Frazzoli, ``Sampling-based algorithms for optimal motion planning,'' \emph{The international journal of robotics research}, vol.~30, no.~7, pp. 846--894, 2011.

\bibitem{pivtoraiko2011kinodynamic}
M.~Pivtoraiko and A.~Kelly, ``Kinodynamic motion planning with state lattice motion primitives,'' in \emph{2011 IEEE/RSJ International Conference on Intelligent Robots and Systems}.\hskip 1em plus 0.5em minus 0.4em\relax IEEE, 2011, pp. 2172--2179.

\bibitem{zucker2013chomp}
M.~Zucker, N.~Ratliff, A.~D. Dragan, M.~Pivtoraiko, M.~Klingensmith, C.~M. Dellin, J.~A. Bagnell, and S.~S. Srinivasa, ``Chomp: Covariant hamiltonian optimization for motion planning,'' \emph{The International Journal of Robotics Research}, vol.~32, no. 9-10, pp. 1164--1193, 2013.

\bibitem{natarajan2021interleaving}
R.~Natarajan, H.~Choset, and M.~Likhachev, ``Interleaving graph search and trajectory optimization for aggressive quadrotor flight,'' \emph{IEEE Robotics and Automation Letters}, vol.~6, no.~3, pp. 5357--5364, 2021.

\bibitem{loui1983optimal}
R.~P. Loui, ``Optimal paths in graphs with stochastic or multidimensional weights,'' \emph{Communications of the ACM}, vol.~26, no.~9, pp. 670--676, 1983.

\bibitem{ren2022enhanced}
Z.~Ren, R.~Zhan, S.~Rathinam, M.~Likhachev, and H.~Choset, ``Enhanced multi-objective a* using balanced binary search trees,'' in \emph{Proceedings of the International Symposium on Combinatorial Search}, vol.~15, no.~1, 2022, pp. 162--170.

\bibitem{HernandezYBZSKS23}
C.~Hern{\'{a}}ndez, W.~Yeoh, J.~A. Baier, H.~Zhang, L.~Suazo, S.~Koenig, and O.~Salzman, ``Simple and efficient bi-objective search algorithms via fast dominance checks,'' \emph{Artif. Intell.}, vol. 314, p. 103807, 2023.

\bibitem{ren22mopbd}
Z.~Ren, S.~Rathinam, M.~Likhachev, and H.~Choset, ``Multi-objective path-based {D}* lite,'' \emph{IEEE Robotics and Automation Letters}, vol.~7, no.~2, pp. 3318--3325, 2022.

\bibitem{ren22mosipp}
------, ``Multi-objective safe-interval path planning with dynamic obstacles,'' \emph{IEEE Robotics and Automation Letters}, vol.~7, no.~3, pp. 8154--8161, 2022.

\bibitem{ehrgott2005multicriteria}
M.~Ehrgott, \emph{Multicriteria optimization}.\hskip 1em plus 0.5em minus 0.4em\relax Springer Science \& Business Media, 2005, vol. 491.

\bibitem{schutze2012using}
O.~Schutze, X.~Esquivel, A.~Lara, and C.~A.~C. Coello, ``Using the averaged hausdorff distance as a performance measure in evolutionary multiobjective optimization,'' \emph{IEEE Transactions on Evolutionary Computation}, vol.~16, no.~4, pp. 504--522, 2012.

\bibitem{hargraves1987direct}
C.~R. Hargraves and S.~W. Paris, ``Direct trajectory optimization using nonlinear programming and collocation,'' \emph{Journal of guidance, control, and dynamics}, vol.~10, no.~4, pp. 338--342, 1987.

\bibitem{underactuated}
\BIBentryALTinterwordspacing
R.~Tedrake, \emph{Underactuated Robotics}, 2022. [Online]. Available: \url{http://underactuated.mit.edu}
\BIBentrySTDinterwordspacing

\bibitem{moore2018opty}
J.~K. Moore and A.~J. van~den Bogert, ``opty: Software for trajectory optimization and parameter identification using direct collocation,'' \emph{Journal of Open Source Software}, 2018.

\bibitem{wachter2006implementation}
A.~W{\"a}chter and L.~T. Biegler, ``On the implementation of an interior-point filter line-search algorithm for large-scale nonlinear programming,'' \emph{Mathematical programming}, vol. 106, no.~1, pp. 25--57, 2006.

\bibitem{jaillet2008transition}
L.~Jaillet, J.~Cort{\'e}s, and T.~Sim{\'e}on, ``Transition-based rrt for path planning in continuous cost spaces,'' in \emph{2008 IEEE/RSJ International Conference on Intelligent Robots and Systems}.\hskip 1em plus 0.5em minus 0.4em\relax IEEE, 2008, pp. 2145--2150.

\bibitem{abbyasov2020automatic}
B.~Abbyasov, R.~Lavrenov, A.~Zakiev, K.~Yakovlev, M.~Svinin, and E.~Magid, ``Automatic tool for gazebo world construction: from a grayscale image to a 3d solid model,'' in \emph{2020 IEEE International Conference on Robotics and Automation (ICRA)}.\hskip 1em plus 0.5em minus 0.4em\relax IEEE, 2020, pp. 7226--7232.

\bibitem{kanayama1990stable}
Y.~Kanayama, Y.~Kimura, F.~Miyazaki, and T.~Noguchi, ``A stable tracking control method for an autonomous mobile robot,'' in \emph{Proceedings., IEEE International Conference on Robotics and Automation}.\hskip 1em plus 0.5em minus 0.4em\relax IEEE, 1990, pp. 384--389.

\bibitem{lee2013hierarchical}
J.~Lee \emph{et~al.}, ``Hierarchical controller for highly dynamic locomotion utilizing pattern modulation and impedance control: Implementation on the mit cheetah robot,'' Ph.D. dissertation, Massachusetts Institute of Technology, 2013.

\end{thebibliography}

\end{document}